\documentclass[10pt,twocolumn,letterpaper]{article}

\usepackage{iccv}
\usepackage{times}
\usepackage{epsfig}
\usepackage{graphicx}
\usepackage{amsmath}
\usepackage{amssymb}
\usepackage{booktabs}
\usepackage{bbding}
\usepackage{bm}
\usepackage{cite}
\usepackage{url}
\usepackage[numbers,sort&compress]{natbib}
\usepackage[accsupp]{axessibility}  

\usepackage[pagebackref=true,breaklinks=true,letterpaper=true,colorlinks,bookmarks=false]{hyperref}

\iccvfinalcopy 

\def\iccvPaperID{****} 

\ificcvfinal\fi

\begin{document}

\title{GIAOTracker: A comprehensive framework for MCMOT with global information and optimizing strategies in VisDrone 2021}

\author{Yunhao Du$^1$, Junfeng Wan$^1$, Yanyun Zhao$^{1,2}$, Binyu Zhang$^1$, Zhihang Tong$^1$, Junhao Dong$^1$ \\
$^1$Beijing University of Posts and Telecommunications \\
$^2$Beijing Key Laboratory of Network System and Network Culture, China\\
{\tt\small \{dyh\_bupt,wanjunfeng,zyy,zhangbinyu,tongzh,djh1999\}@bupt.edu.cn}

}

\maketitle
\ificcvfinal\thispagestyle{empty}\fi

\begin{abstract}
   In recent years, algorithms for multiple object tracking tasks have benefited from great progresses in deep models and video quality. 
   However, in challenging scenarios like drone videos, they still suffer from problems, such as small objects, camera movements and view changes. 
   In this paper, we propose a new multiple object tracker, which employs \textbf{G}lobal \textbf{I}nformation \textbf{A}nd some \textbf{O}ptimizing strategies, named GIAOTracker. 
   It consists of three stages, i.e., online tracking, global link and post-processing. 
   Given detections in every frame, the first stage generates reliable tracklets using information of camera motion, object motion and object appearance. 
   Then they are associated into trajectories by exploiting global clues and refined through four post-processing methods. 
   With the effectiveness of the three stages, GIAOTracker achieves state-of-the-art performance on the VisDrone MOT dataset and wins the 2nd place in the VisDrone2021 MOT Challenge.
\end{abstract}


\section{Introduction}
   Drones (or general UAVs) equipped with cameras have been widely applied to various fields, e.g., agriculture, meteorology, aerial photography, fast delivery and surveillance ~\cite{visdrone}.
   Consequently, drone video understanding is receiving increasingly attention. 
   Multiple Object Tracking (MOT), which aims to identify and track one or multiple categories of objects, is the key component in autonomous drone systems. 
   However, it suffers from problems, including large number of small objects by aerial capturing, irregular object motion, camera movements, variant views, occlusion from trees and bridges, etc. 
   It is more complex and difficult to solve these problems than those in camera fixed scenarios, e.g., surveillance videos, which makes MOT in drone videos still a challenging task.

   In this paper, we present a comprehensive framework with global information and optimizing strategies for Multi-Class Multi-Object Tracking (MCMOT) in drone videos, named GIAOTracker. 
   To alleviate detection noises, a new feature storage and update strategy EMA Bank is proposed to maintain both variant feature states and information of feature changes simultaneously. 
   As for object motion modeling, linear Kalman filter algorithm is widely used \cite{wojke2017simple,bewley2016simple,zhang2020fairmot,wang2020towards,pan2019multi}, which simply sets a uniform measurement noise scale to all objects without considering detection quality.
   To obtain more accurate motion state, we propose a Noise Scale Adaptive Kalman algorithm (NSA Kalman) which adaptively modulates the noise scale according to the quality of object detection.

   Trajectory global information plays an important role in solving the fracture problem. 
   Global information is not well exploited in many recent MOT works \cite{wojke2017simple,bewley2016simple,xiang2015learning,zhang2020fairmot,wang2020towards,bochinski2017high,bochinski2018extending}. 
   Instead, we introduce a global link stage to associate tracklets into trajectories. 
   Specifically, to reduce noises caused by occlusion and view changes, we propose a novel tracklet appearance feature extractor GIModel (\textbf{G}lobal \textbf{I}nformation model), 
   which extracts both global and part spatial features in each frame and fuses them with a self-attention based temporal modeling block for more robust representation.

   For MCMOT task in complex scenes, e.g., UAV videos, reasonable post-processing strategies could greatly improve the tracking performance. 
   However, only a few works focus on the post-processing procedure \cite{cofe,chu2021spatial,peng2020tpm}. 
   To refine tracking results in a more comprehensive way, we propose to use four post-processing methods.
   To remove redundant trajectories caused by duplicate detections, a temporal-IoU based NMS (Non-Maximum Suppression) between two trajectories is used. 
   Then, missing detections are interpolated linearly into the trajectory gaps as in TPM \cite{peng2020tpm}. 
   Considering that longer trajectories tend to be more accurate, we use a length-dependent coefficient to rescore trajectories frame by frame. 
   Last but not least, to the best of our knowledge, few works explore the fusion strategy for the MOT task. 
   Inspired by SoftNMS \cite{bodla2017soft}, we introduce TrackNMS to fuse different tracking results, which significantly improves tracking performance.

   On the VisDrone MOT test-challenge dataset, GIAOTracker achieves \textbf{52.55} mAP with detections generated by DetectoRS \cite{qiao2021detectors}. 
   After fusing two tracking results (46.66 mAP \& 52.55 mAP), we achieve \textbf{54.18} mAP and win the 2nd place in the VisDrone2021 MOT Challenge. 
   Furthermore, with annotation detections as the input, our tracking performance improves by 92.4\% (from 43.12 to 82.97 mAP) on the test-dev dataset, 
   which proves the effectiveness of our tracking framework.

   The main contributions of this article are summarized as follows:
   \begin{itemize}
      \setlength{\itemsep}{0pt}
      \item For MCMOT in drone videos, we present a comprehensive framework with global information and some optimizing strategies (GIAOTracker), which consists of three stages, i.e., online tracking, global link and post-processing.
      \item We propose EMA Bank strategy and NSA Kalman algorithm, which aim at more accurate and robust association.
      \item We introduce a tracklet feature extractor GIModel, which extracts frame-level global-part features and then fuses them with a self-attention based temporal modeling block.
      \item We explore a series of reasonable and effective post-processing strategies, including trajectory denoising, detection box interpolation, trajectory re-scoring and tracking model fusion.
   \end{itemize}

\section{Related works}

\subsection{SDE and JDE}
   Most recent MOT methods could be classified into two categories: “Separate Detection and Embedding (SDE)” and “Joint Detection and Embedding (JDE)” \cite{wang2020towards}. 
   SDE, which is also termed as tracking-by-detection \cite{wojke2017simple,bewley2016simple,fagot2016improving,keuper2018motion,milan2015multi,pirsiavash2011globally,wang2015tracking,xiang2015learning}, consists of the following two steps: 
   1) detection, in which all objects are localized and classified in sequences \cite{redmon2016you,redmon2017yolo9000,redmon2018yolov3,bochkovskiy2020yolov4,wang2021scaled,yolov5,girshick2014rich,girshick2015fast,ren2015faster,cai2018cascade}; 
   2) association, where detections belonging to the same object are associated by assigning the same ID \cite{fang2018recurrent,huang2008robust,dehghan2015target,zamir2012gmcp,zhang2008global,chen2017enhancing,kim2015multiple,milan2017online,ondruska2016deep,chu2019famnet,braso2020learning,xu2020train}. 
   SDE strategy optimizes detection and embedding separately, which is more flexible and suitable for complex scenarios. 
   However, it tends to cost much time in inference. Instead, JDE incorporates detection model and embedding into a unified framework \cite{zhang2020fairmot,wang2020towards,voigtlaender2019mots,kim2016cdt,feichtenhofer2017detect,bergmann2019tracking,zhou2020tracking,peng2020chained,pang2020tubetk,sun2020simultaneous}. 
   It typically modifies detectors, e.g., Faster R-CNN \cite{ren2015faster}, CenterNet \cite{zhou2019objects}, YOLOv3 \cite{redmon2018yolov3} by adding a predictor \cite{bergmann2019tracking,zhou2020tracking,han2020mat,zhang2020multiple} or an embedding branch \cite{zhang2020fairmot,wang2020towards} 
   and leverages them to implement detection and tracking jointly. Generally, JDE methods usually behave better and faster than SDE in common applications. 
   Whereas, they would fail when applied to more sophisticated scenarios.
   
   In this paper, in light of the complexity of drone videos, our GIAOTracker follows the SDE paradigm. 
   It allows us to train the detector independently, which could generate more accurate localization and classification results than JDE paradigm.

\subsection{Online and Offline}
   We could also divide MOT methods into online tracking and offline tracking methods on whether using global information. 
   Online methods perform association on the fly without knowledge of future information \cite{wojke2017simple,bewley2016simple,xiang2015learning,zhang2020fairmot,wang2020towards,bochinski2017high,bochinski2018extending}. 
   Most recent MOT methods are online and achieve state-of-the-art performance. 
   Besides, compared with offline methods, online tracking has more application scenarios, e.g., real-time tracking system. 
   Offline methods, on the other hand, are allowed to employ future frames and tend to result in better tracking quality \cite{wang2019exploit,girbau2021multiple,li2021semi,pang2021quasi,wang2021split,hornakova2020lifted}.

   Inspired by the hierarchical data association strategy \cite{huang2008robust,yang2012multi}, our GIAOTracker includes three stages: online tracking, global link and post-processing. 
   The first stage (GIAOTracker-Online) performs online tracking to generate reliable tracklets and the second stage associates them into trajectories with global information. 
   This hierarchical framework gives a tradeoff between accuracy and flexibility, as one could only apply GIAOTracker-Online for those online tracking scenarios, 
   or use the full GIAOTracker method for better performance. 
   Moreover, the global link and post-processing stages are both plug-and-play, which can be plugged into any existing MOT frameworks easily.

\begin{figure*}[htbp]
   \centering
   \includegraphics[width = 0.85\textwidth]{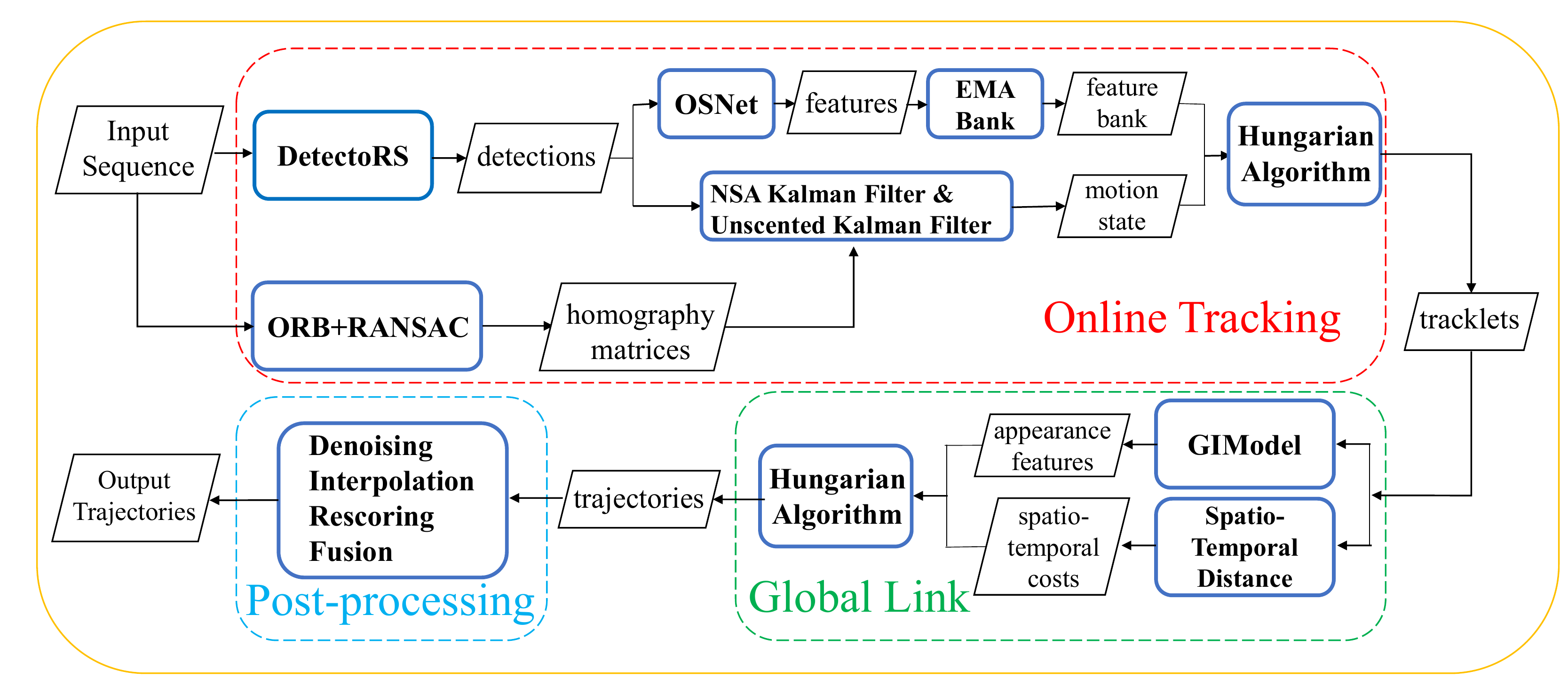}
   \caption{Overview of our proposed GIAOTracker pipeline for MCMOT}
   \end{figure*}

\subsection{MOT in Drone}
   MOT in drone videos is a challenging task due to small objects, camera movements, variant views, etc. 
   VisDrone2018 dataset is proposed in \cite{zhu2018vision}, which focuses on core problems in computer vision.
   VisDrone-VDT2018 \cite{zhu2018visdrone}, VisDrone-MOT2019 \cite{wen2019visdrone} and VisDrone-MOT2020 \cite{fan2020visdrone} 
   propose abundant methods which greatly improve the ability of intelligent system to understand drone videos.

   V-IOU Tracker \cite{bochinski2018extending} improves IOU Tracker \cite{bochinski2017high} by visual tracking \cite{henriques2014high,kalal2010forward} to continue a track if no detection is available, 
   which achieves state-of-the-art performance at high processing speeds in VisDrone-VDT2018 Challenge. 
   However, it doesn’t take camera movements into account and global information is not well exploited. 
   Thus, we use ORB \cite{rublee2011orb} and RANSAC \cite{fischler1981random} to deal with camera movements and leverage a novel tracklet feature extractor GIModel to implement associations among tracklets. 
   HMTT \cite{pan2019multi} provides a hierarchical multi-target tracker based on detection for drone vision, 
   where four stages are proposed to deal with multiple problems like variant views, unreliable detections and missing detections. 
   But it pays little attention on post-processing. In contrast, we highlight the importance of post-processing procedure and apply four methods to refine tracking results. 
   COFE \cite{cofe} proposes a coarse-to-fine tracking framework to reduce the classification noises and wins the first place in VisDrone-MOT2020 Challenge. 
   We improve it by replacing the “hard-vote” mechanism with “soft-vote” and employing global information to further improve tracking performance.

\section{Method}
   We aim at MCMOT in drone videos in a hierarchical data association way. 
   Figure 1 illustrates our GIAOTracker framework built upon SDE paradigm, which consists of three stages, i.e., online tracking, global link and post-processing.

\subsection{Online Tracking}

   \begin{figure*}[htbp]
      \centering
      \includegraphics[width =0.8\textwidth]{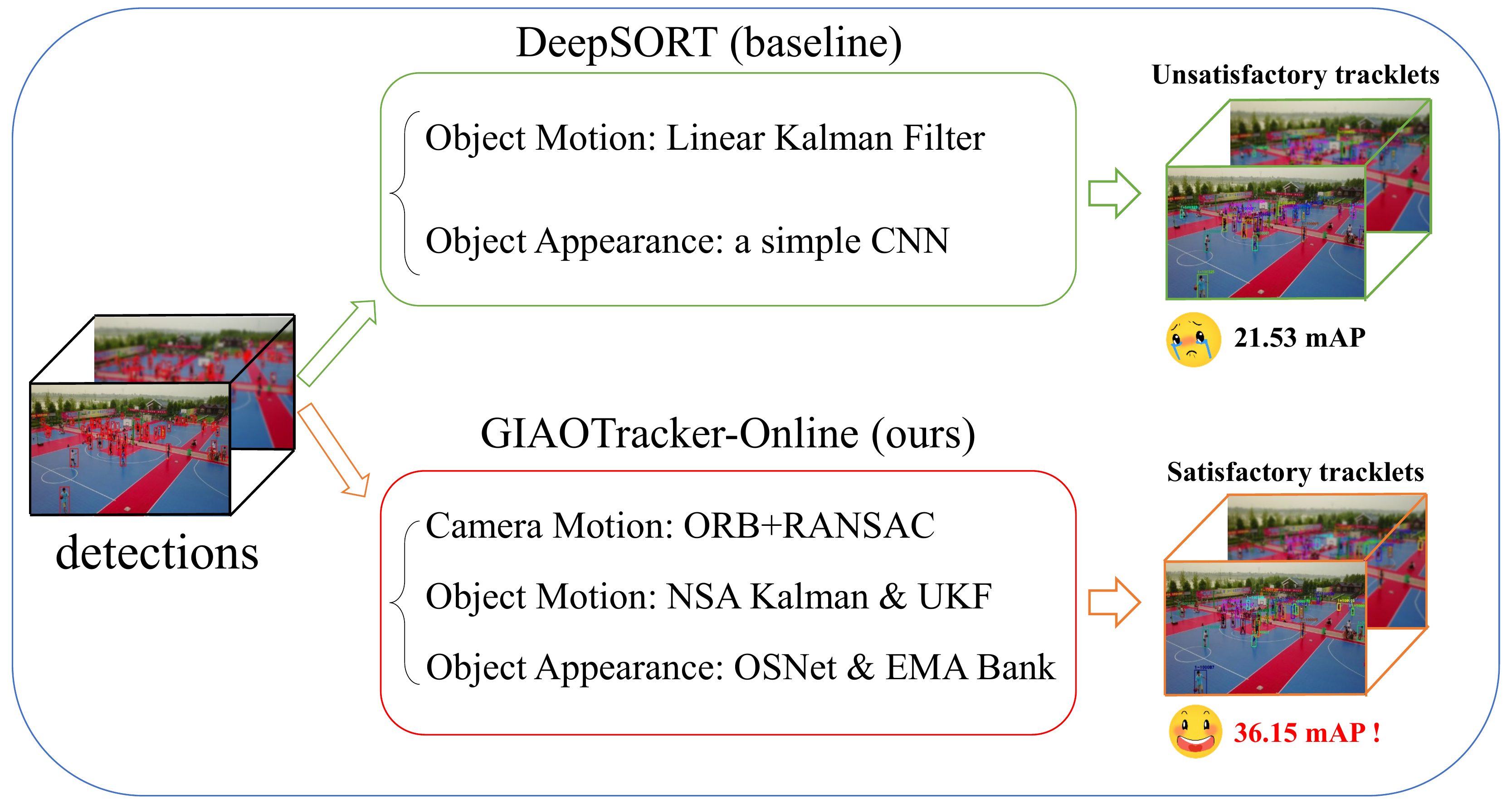}
      \caption{Comparision between DeepSORT (baseline) and GIAOTracker-Online (ours).}
   \end{figure*}

   DeepSORT \cite{wojke2017simple} is a typical and strong MOT algorithm following tracking-by-detection paradigm, 
   which generates detections first and then associates them with object motion information and appearance features (top in Figure 2). 
   Particularly, all modules in DeepSORT, i.e. detector, Kalman and feature extractor, are plug-and-play, which allows flexibility to improve it. 
   Consequently, the online tracking stage of our GIAOTracker applies DeepSORT as baseline. Figure 2 shows the comparison between DeepSORT and our GIAOTracker-Online. 
   Taking sequences as input, we use DetectoRS \cite{qiao2021detectors} to generate detections $\{b^t\}_{t=1}^T$ frame by frame and then link them into tracklets $\{tl_n\}_{n=1}^N$. 
   To deal with camera movement, we utilize ORB \cite{rublee2011orb} and RANSAC \cite{fischler1981random} to fastly align inter-frame images. 
   Then we improve the baseline from two aspects, i.e., object appearance and object motion. 
   In order to obtain more robust appearance features, we replace the simple feature extractor in DeepSORT with OSNet \cite{zhou2019omni} and train it on VisDrone dataset. 
   A new feature storage and update strategy EMA Bank is proposed to achieve more accurate association between tracklets and detections.

   \noindent \textbf{EMA Bank.} There exist two mainstream feature storage and update approaches. DeepSORT \cite{wojke2017simple} implements a feature bank to store raw features of previous $L_b$ detections 
   and use them to calculate the minimum cosine distance with detection features.
   For tracklet $tl_i$, its feature bank is $FB_i = \{\bm{f^t_i}\}_{t=1}^T$, where $\bm{f_i^t}$ is the raw detection feature.
   Such mechanism maintains variant feature states of one tracklet and is robust to sudden changes of object appearance. 
   However, simply using raw features is sensitive to detection noises. 
   Instead, JDE/FairMOT \cite{wang2020towards,zhang2020fairmot} use an EMA (Exponential Moving Average) feature update strategy, in which only one feature is maintained for every tracklet. 
   For tracklet $tl_i$, its feature state $\bm{e_i}$ is updated by:
   
   \begin{equation}
      \bm{e_i^t}=\alpha \bm{e_i^{t-1}}+(1-\alpha) \bm{f_i^t} \label{XX}
   \end{equation}
   
   \noindent where $\bm{f_i^t}$ is the appearance embedding of the detection in frame $t$ and $\alpha$ is the momentum term.
   This incremental feature update strategy leverages the information of inter-frame feature changes and could depress detection noises. 
   To integrate the advantages of both approaches above, we explore an intuitive method, named EMA Bank. 
   For tracklet $tl_i$, its bank is $EB_i = \{\bm{e_i^t}\}_{t=1}^T$, where $\bm{e_i^t}$ is calculated by equation (1). 
   The proposed EMA Bank simultaneously takes multi-frame information and inter-frame change information into account, which is more suitable for complex scenarios.

   In online tracking framework, motion prediction is another key module, in which Kalman filter \cite{kalman1960new} is commonly used. 
   For vehicle objects, we apply Unscented Kalman Filter (UKF) algorithm \cite{wan2000unscented} which is more robust for nonlinear motion. 
   We also propose a modified Kalman filter algorithm named NSA Kalman Filter, which could adaptively modulate the noise scale during the state update procedure.

   \begin{figure}[htbp]
      \centering
      \includegraphics[width = .45\textwidth]{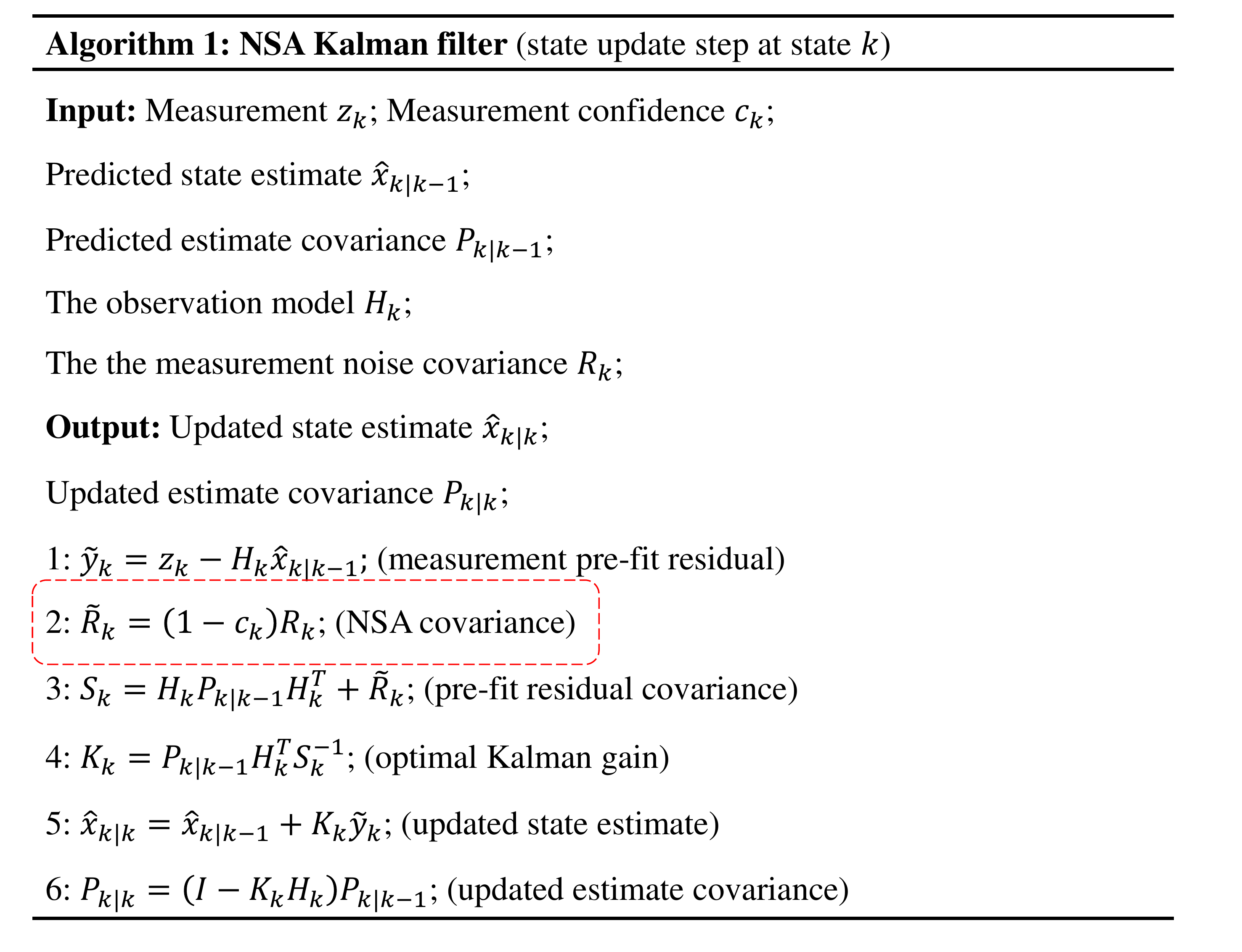}
   \end{figure}

   \noindent \textbf{NSA Kalman.} In DeepSORT, Kalman filter based on linear motion hypothesis is used to model objects motion. 
   It consists of state estimation step and state update step. 
   In the first step, Kalman filter produces estimates of current state variables, along with their uncertainties. 
   Then these estimates are updated with a weighted average of the estimated state and the measurement. 
   Specifically, it uses the measurement noise covariance $R_*$ to represent the measurement (i.e., detections in the current frame) noise scale. 
   A larger noise scale means a smaller weight of the measurement during state update step, since its larger uncertainty. 
   In Kalman algorithm \cite{kalman1960new}, the noise scale is a constant matrix. 
   However, intuitively different measurement contains different scales of noise. 
   In substance, measurement noise scale should vary with detection confidence. 
   Therefore, we propose a formula to adaptively calculate the noise covariance, named NSA noise covariance $\tilde R_k$:

   \begin{equation}
      \tilde R_k = (1 - c_k) R_k \label{a}
   \end{equation}

   \noindent where $R_k$ is the preset constant measurement noise covariance and $c_k$ is the detection confidence score at state $k$.
   The whole state update of our NSA Kalman filter is shown in the Algorithm 1, where the NAS step is marked with a red dotted box. 
   Experimental results show that it significantly improves the tracking performance, though our NAS Kalman is simple (Table 1).

   \noindent \textbf{Rough2Fine.} We adopt soft voting method to classify tracklets for GIAOTracker-Online. 
   It’s well known that some object categories are difficult to distinguish in VisDrone, e.g., car \& van. 
   Therefore, instead of tracking different categories independently, we follow the Coarse-to-Fine pipeline in COFE \cite{cofe} and improve it with a “soft-vote” mechanism (named Rough2Fine). 
   As in COFE, we first implement rough-class tracking and then determine fine classes of trajectories with voting mechanism. 
   Different from the “hard-vote” in COFE, our “soft-vote” mechanism assigns multiple classes to a single trajectory, 
   in which the voting weight is positively correlated with confidence score. 
   Experiments show that “soft-vote” is more robust to classification errors than “hard-vote” (Table 1). 

   Based on the above camera correction, object motion prediction and appearance feature processing strategies, 
   we associate object detections to form reliable tracklets in an online tracking way. 
   Though it has improved the baseline by a large margin, we argue that future information is not exploited by our GIAOTracker-Online, so its performance is still limited. 
   In the next section, we’ll introduce a global link algorithm which links tracklets into trajectories by employing global information.

   \begin{figure*}[htbp]
      \centering
      \includegraphics[width = 0.9\textwidth]{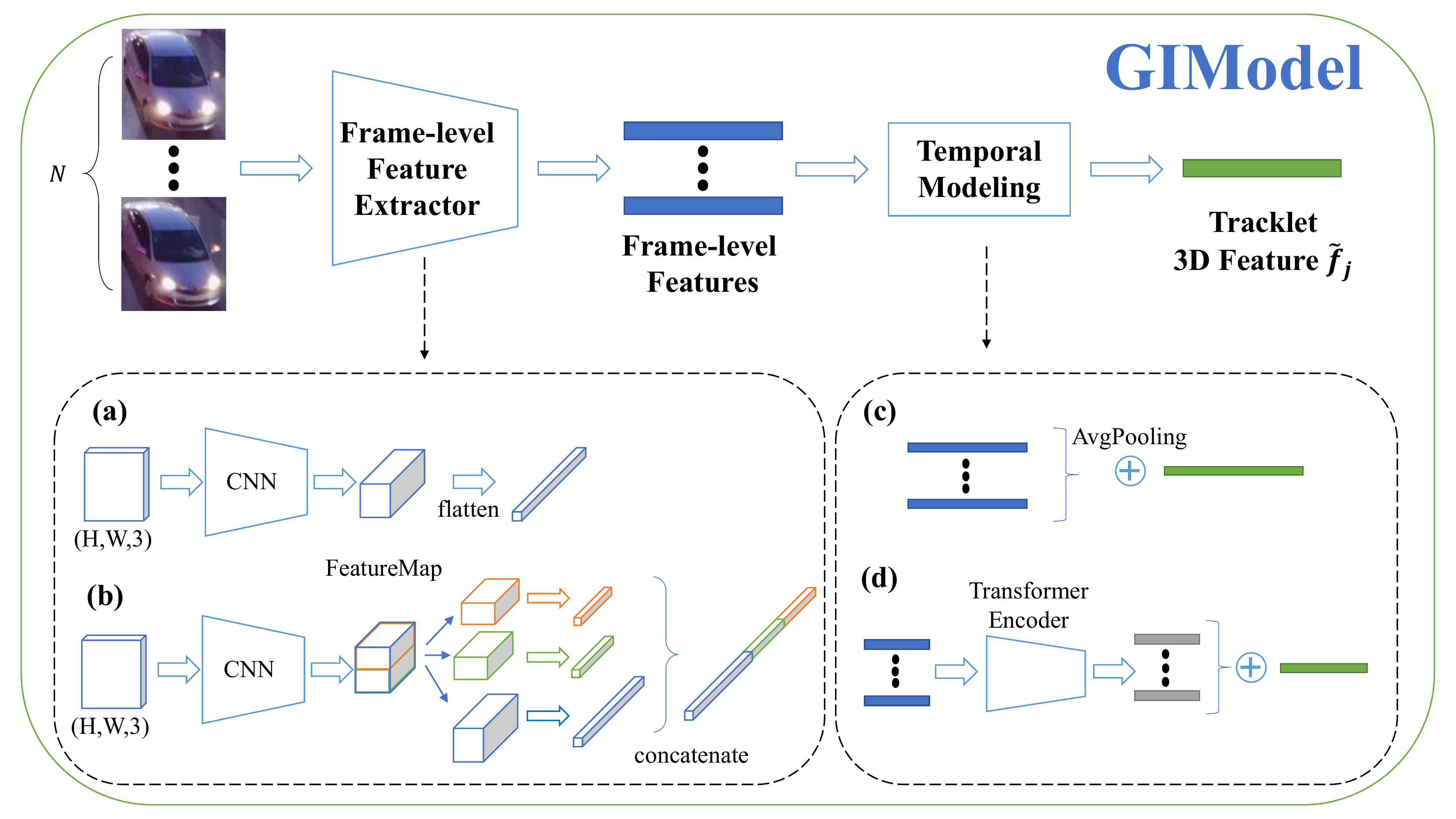}
      \caption{The framework of our GIModel. 
      (a) Simply extracting global features based on a CNN. 
      (b) Given CNN feature maps, extracting both global and part features. 
      (c) Simply averaging frame-level features for temporal modeling. 
      (d) Temporal modeling with a Transformer encoder layer. We take ResNet50-TP \cite{gao2018revisiting} as baseline and imporve it by adding part-level features ((b) vs. (a)) and self-attention based temporal modeling ((d) vs. (c)).}
   \end{figure*}

\subsection{Global link}
   Global link stage links short tracklets to long trajectories based on Hungarian algorithm \cite{kuhn1955hungarian}. 
   In order to make full use of global information for tracklet association, we integrate appearance and spatio-temporal distances of tracklets into a single matching cost for Hungarian algorithm. 
   In this section, we first introduce our appearance feature extractor GIModel for tracklets, and then present the tracklet association algorithm.

   \noindent \textbf{GIModel.} We build our GIModel based on ResNet50-TP \cite{gao2018revisiting} and improve it by adding part-level features and self-attention based temporal modeling \cite{vaswani2017attention}. 
   Figure 3 shows the framework of our GIModel. 
   Taking consecutive $N$ frames of a tracklet clip as input, GIModel extracts frame-level features first and outputs the clip features by temporal modeling. 
   Different from the baseline who only extracts spatial global features (Figure 3 (a)), we add part-level features supervised by additional triplet loss \cite{hermans2017defense} as shown in Figure 3 (b).
   This enables GIModel to focus on detailed features of different parts of objects and be more robust to occlusion. 
   In inference, global and part features are directly concatenated. 
   As for temporal modeling, instead of simply fusing frame-level features by average pooling (Fig. 2 (c)), 
   we implement inter-frame information interaction with a Transformer encoder layer \cite{vaswani2017attention} showed by Figure 3 (d) in order to aggregate features of multi-frames and suppress noises for one clip.
   Given $N$ frames of clip $j$, the self-attention based features $\{\bm{\hat f_j^t}\}_{n=1}^N$ are obtained by a Transformer encoder layer:
   \begin{equation}
      \{\bm{\hat f_j^t}\}_{n=1}^N = TEL(\{\bm{f_j^t}\}_{n=1}^N) \label{b}
   \end{equation}
   \noindent where $\bm{f_j^t}$ represents the raw frame-level features and $TEL(\cdot)$ represents the Transformer encoder layer. Then the 3D feature $\bm{\tilde f_j}$ for clip $j$ is calculated by:
   \begin{equation}
      \bm{\tilde f_j} = {1 \over N} \sum_{n=1}^N \bm{\hat f_j^n} \label{c}
   \end{equation}
   \noindent The feature bank for tracklet $tl_i$ is the set of its clip features $\tilde F_i = \{\bm{\tilde f_j^i}\}_{j=1}^M$.

   To conclude, part features focus on detailed spatial information and self-attention based temporal modeling aggregates temporal context information more effectively. 
   Experimental results (Table 2) show that our GIModel performs better than baseline by a large margin.

   \noindent \textbf{Association.} We calculate the matching cost matrix used by Hungarian algorithm with appearance cost and spatio-temporal cost for tracklet association.
   Specifically, given $tl_i$ and $tl_j$, their appearance cost is the smallest cosine distance between their feature banks $\tilde F_i$ and $\tilde F_j$:
   \begin{equation}
      C_a(i,j) = min\{1-\bm{{\tilde f_i^n}^T} \bm{\tilde f_j^m} | \bm{\tilde f_i^n} \in \tilde F_i, \bm{\tilde f_j^m} \in \tilde F_j\} \label{d}
   \end{equation}
   Additionally, the spatio-temporal distance costs $C_s(i,j)$ and $C_t(i,j)$ measure the time interval and space distances for two tracklets respectively. 
   If the cost meets the threshold constaint,
   \begin{equation}
      C_a(i,j) < Th_a \ and \ C_t(i,j) < Th_t \ and \ C_s(i,j) < Th_s \label{e}
   \end{equation}
   \noindent then matching cost is calculated as follows:
   \begin{equation}
      C(i,j) = \lambda_a C_a(i,j) + \lambda_t C_t(i,j) + \lambda_s C_s(i,j) \label{f}
   \end{equation}
   \noindent where $Th_a, Th_t, Th_s$ are the preset thresholds of appearance feature cost, time cost and space  cost respectively,
   and $\lambda_a, \lambda_t, \lambda_s$ are weight coefficients.
   Cosine distance is an effective metric for measuring appearance features. 
   Time interval and space distances of tracklets are directly used to avoid errors in velocity estimation for object motion. 
   In this way, based on Hungarian algorithm, we associate the tracklets to form long trajectories well.

   In this section, we introduce a tracklet association algorithm based on tracklets appearance features and spatio-temporal distances. 
   Particularly, GIModel is proposed to extract representative 3D features for tracklets, which is robust to detection noises and abrupt changes of object appearance. 
   Next, we’ll present some post-processing methods to further improve tracking accuracy.

\subsection{Post-processing}
   In this section, we explore four post-processing procedures to refine the trajectories, i.e., denoising, interpolation, rescoring and fusion.

   \noindent \textbf{Denoising.} There exist some duplicate detections which would result in redundant trajectories. 
   STGT \cite{chu2021spatial} uses detection-wise spatio-IoU based matching procedure to remove unmatched detection candidates. 
   Instead, our denoising algorithm uses trajectory-wise temporal-IoU to implement SoftNMS \cite{bodla2017soft} between trajectories.

   \noindent \textbf{Interpolation.} Within one trajectory, missing detections would also decrease tracking accuracy. 
   As in TPM \cite{peng2020tpm}, detections are interpolated linearly into gaps of the trajectory. 
   Considering that larger gaps will bring more noises, only missing frames less than 60 are filled in.

   \noindent \textbf{Rescoring.} While evaluating, the average score is used to measure the quality of trajectories. 
   According to our observation, however, longer trajectories tend to be more accurate. 
   Therefore, we use a length-dependent coefficient to rescore trajectories per frame.
   For trajectory $i$ whose length is $l_i$, the rescoring weight is calculated as:
   \begin{equation}
      \omega_i = {{1 - e^{-l_i / \tau}} \over {1 + e^{-l_i / \tau}}} \label{g}
   \end{equation}
   \noindent where the temperature factor $\tau = 25$. For frame $j$, its confidence $s_i^j$ is rescored as $\hat s_i^j = \omega_i \cdot s_i^j$.
   In this way, the confidence of long trajectories is relatively increased while decreased of short trajectories.

   \noindent \textbf{Fusion.} Model fusion is a common strategy to improve performance in some computer vision tasks, such as image classification \cite{dietterich2000ensemble} and object detection \cite{solovyev2021weighted}.
   To the best of our knowledge, few works focus on fusion strategies for MOT task. 
   Inspired by the success of NMS-based model fusion on object detection task, we propose TrackNMS to fuse different tracking results. 
   In short, TrackNMS is based on the idea of SoftNMS \cite{bodla2017soft} with two main differences:
   \begin{enumerate}
      \item [1.] IoU: SoftNMS uses spatio-IoU of detections to determine the degree of suppression. Instead, TrackNMS uses temporal-IoU between trajectories.
      \item [2.] Sort: Unlike SoftNMS who sorts detections by their scores, TrackNMS applies the sum (instead of “mean”) of trajectory frame-level scores as the sorting basis, 
      which means longer trajectories tend to have higher priority.
   \end{enumerate}
   \noindent Experimental results (Table 4) show that our TrackNMS works well. 

\section{Experiment}

\subsection{Dataset and Metrics}

   \begin{table*}[h]
      \begin{center}
      \begin{tabular}{cc|c|c|c|c|c|c|c}
      \toprule[2pt]
         & \textbf{Method} & \textbf{add} &\textbf{mAP} &\textbf{mAP-ped.} &\textbf{mAP-car} &\textbf{mAP-van} &\textbf{mAP-truck} &\textbf{mAP-bus} \\ \midrule[1pt]
         & baseline \cite{wojke2017simple} & - & 21.53 & 14.21 & 30.37 & 20.34 & 22.22 & 20.49 \\
         & +ORB & \checkmark & 29.69 & 14.32 & 51.68 & 26.97 & 25.37 & 30.14 \\
         & +EMA Bank & \checkmark & 30.06 & 14.77 & 51.14 & 27.45 & 26.78 & 30.14 \\
         & +NSA Kalman & \checkmark & 32.30 & 18.59 & 54.84 & 27.55 & 30.37 & 30.14 \\
         & +OSNet & \checkmark  & 32.66 & \textbf{19.61} & 55.74 & 27.68 & 30.12 & 30.14 \\
         & +R2F-hard & ×  & 32.29 & 19.39 & 55.83 & 30.18 & 29.61 & 26.46 \\
         & +R2F-soft & \checkmark & 33.87 & 19.60 & 55.93 & \textbf{30.83} & 29.29 & 33.67 \\
         & +UKF & \checkmark  & \textbf{36.15} & 19.60 & \textbf{57.61} & 30.80 & \textbf{33.00} & \textbf{39.76} \\
      \bottomrule[2pt]
      \end{tabular}
      \end{center}
      \caption{Ablation studies of the online tracking stage. “\checkmark” in the “add” line means the corresponding component is added to GIAOTracker. 
      “R2F-hard” means “Rough2Fine” using “hard-vote” strategy and “R2F-soft” uses “soft-vote” (best in bold).}
   \end{table*}

   \begin{table}[h]
      \begin{center}
      \resizebox{0.45\textwidth}{12mm}{
      \begin{tabular}{cc|c|c|c|c}
      \toprule[2pt]
         & \textbf{Method} &\textbf{mAP-P} &\textbf{Rank1-P} &\textbf{mAP-V} &\textbf{Rank1-V} \\ \midrule[1pt]
         & baseline \cite{wojke2017simple} & 52.1 & 75.5 & 65.4 & 90.1 \\
         & +train & 76.4 & 89.7 & 80.7 & 94.0 \\  
         & +sat & 80.3 & 91.8 & 83.8 & \textbf{95.9} \\
         & +part & \textbf{81.3} & \textbf{92.2} & \textbf{85.1} & 95.8 \\
      \bottomrule[2pt]
      \end{tabular}}
      \end{center}
      \caption{Effects of different strategies for GIModel, which use ResNet50-TP \cite{gao2018revisiting} pretrained on ImageNet as baseline. 
      “+train” means training it on the VisDrone2021 dataset. 
      “+sat” represents implementing our self-attention based temporal modeling block. 
      “+part” means adding part-level features supervised by triplet loss \cite{hermans2017defense} (best in bold).}
      \end{table}

   The VisDrone MOT dataset consists of 96 sequences (39,988 frames in total).
   Each frame is manually labeled in high quality. Participants are asked to submit tracking results for 5 selected object classes in this challenge, 
   including pedestrian, car, van, bus and truck. 
   Unless otherwise stated, we use both training and validation dataset for training and use test-dev for validation.

   The VisDrone2021 MOT Challenge uses the protocol in \cite{lsvrc2017} to evaluate the tracking performance. 
   Given tracking results which consist of a list of detections with confidence scores and corresponding identities, 
   trajectories are sorted according to the average confidence of their detections. 
   A trajectory is considered correct if the IoU overlap with the ground truth trajectory is larger than a threshold. 
   The final ranking metric is calculated by averaging the mean average precision (mAP) per object class over different thresholds.

\subsection{Implementation details}

   \noindent \textbf{Detection.} We fine-tune the ResNet50-based \cite{he2016deep} DetectoRS \cite{qiao2021detectors} detector on the VisDrone MOT train+val dataset, which is pre-trained on MS COCO \cite{lin2014microsoft} dataset. 
   To avoid overfitting, 1 frame in every 5 frames is sampled while training. 
   Training input scale is set to [1333, 800]. When testing, SoftNMS \cite{bodla2017soft} and multi-scale testing [(1333, 800), (2000, 1200)] are used. 
   It achieves 56.9 AP50 on the test-dev dataset, named DetV1. As the improvement, when training, we cut an image into 4 non-overlapping regions as input and use a larger input scale [1600, 1050].
   When testing, we still use the entire image as input and change the input scale to [(3000, 1969), (3200, 2100), (3400, 2231), (3600, 2362), (3800, 2494), (4000, 2625)]. 
   This improved method achieves 59.4 AP50. Then we fuse it with DetV1 using SoftNMS, whose AP50 is 63.2 on test-dev, named DetV2.

   \noindent \textbf{OSNet.} For training and evaluating the OSNet, we create a ReID dataset based on VisDrone MOT dataset. 
   The frame sampling rate is 5 and the minimum height/width is 32. Any object with an occlusion ratio or truncation ratio greater than 50\% will be removed. 
   The ratio of gallery to probe is set to 7:1. We use the ImageNet \cite{deng2009imagenet} pre-trained model and fine-tune it on our dataset. 
   The input scale is set to [128, 256] for vehicles (i.e., car, van, truck and bus) and [256, 128] for pedestrian.

   \noindent \textbf{GIModel.} Similar to the ReID dataset mentioned above, we also create a VideoReID-like \cite{zheng2016mars,wu2018exploit,li2019global} dataset for GIModel. 
   The main differences are the sampling rate is set to 1 and the ratio of gallery to probe is set to 3:1. 
   The input size is set to [224, 224, 4] for vehicle tracklet clips and [224, 112, 4] for pedestrian. 
   ImageNet \cite{deng2009imagenet} pre-triained model is also used. GIModel is evaluated by mAP and Rank1, which are generally used in ReID and VideoReID tasks \cite{sun2018beyond,zhou2019omni,gao2018revisiting}.

   \noindent \textbf{Others.} In EMA Bank, $L_b = 100$, $\alpha=0.9$ as in DeepSORT \cite{wojke2017simple} and JDE \cite{wang2020towards}. 
   In the global link stage, $Th_a=0.4$, $Th_t=200$, $Th_s=150$ and $\lambda_a=40$, $\lambda_t=1$, $\lambda_s=1$.

\subsection{Ablation}

   In this section, we describe the ablation results on VisDrone MOT test-dev dataset.

   \noindent \textbf{Online Tracking.} In order to evaluate the effect of different components of our online tracking stage, we compare them with baseline \cite{wojke2017simple} as shown in table 1. 
   We have the following four observations: 1) Images matching based on ORB significantly improves the tracking performance which compensates for the camera movements. 
   2) Both stronger feature extractor OSNet and more robust feature storage strategy EMA Bank benefit tracking results. 
   3) NSA Kalman and UKF are far superior to linear Kalman filter algorithm. 
   4) As for Rough2Fine strategy, “soft-vote” exceeds “hard-vote” by a large margin.

   \noindent \textbf{GIModel.} Table 2 presents the effects of different strategies evaluated on our VideoReID dataset for GIModel, which takes ResNet50-TP \cite{gao2018revisiting} as baseline
   (“P” for people, “V” for vehicle).
   Results show that both our self-attention based temporal modeling block and part-level features supervised by triplet loss \cite{hermans2017defense} make GIModel perform better than baseline significantly.

   \noindent \textbf{Post-processing.} We explore the influence of different post-processing procedures in table 3. 
   Denoising and interpolation brings 0.14 mAP and 0.23 mAP gains respectively. 
   We argue that interpolation is complementary to denoising, i.e., denoising removes redundant trajectories caused by replicate detections and interpolation restores missing detections. 
   Furthermore, our rescoring method increases tracking accuracy by 1.64 mAP, which indicates the quality of trajectories is strongly correlated to their length.

   \begin{table*}[h]
      \begin{center}
      \begin{tabular}{cc|c|c|c|c|c|c}
      \toprule[2pt]
         & \textbf{Method} &\textbf{mAP} &\textbf{mAP-ped.} &\textbf{mAP-car} &\textbf{mAP-van} &\textbf{mAP-truck} &\textbf{mAP-bus}\\ \midrule[1pt]
         & GIAOTracker-Global & 38.71 & 23.63 & \textbf{57.95} & 35.19 & 36.35 & 40.43 \\
         & +denoising & 38.85 & 23.90 & 57.82 & 35.51 & 36.59 & 40.43 \\  
         & +interpolation & 39.08 & 25.49 & 57.57 & 35.48 & \textbf{37.21} & 39.63 \\
         & +rescoring & \textbf{40.72} & \textbf{27.10} & 57.46 & \textbf{36.65} & 36.72 & \textbf{45.69} \\
      \bottomrule[2pt]
      \end{tabular}
      \end{center}
      \caption{The influence of different post-processing methods (best in bold).}
   \end{table*}

   \begin{table*}[h]
      \begin{center}
      \begin{tabular}{cc|c|c|c|c|c|c}
      \toprule[2pt]
         & \textbf{Method} &\textbf{mAP} &\textbf{mAP-1} &\textbf{mAP-4} &\textbf{mAP-5} &\textbf{mAP-6} &\textbf{mAP-9} \\ \midrule[1pt]
         & baseline \cite{wojke2017simple} & 21.53 & 14.21 & 30.37 & 20.34 & 22.22 & 20.49 \\
         & GIAOTracker-Online & 36.15 & 19.60 & 57.61 & 30.80 & 33.00 & 39.76 \\
         & GIAOTracker-Global & 38.71 & 23.63 & 57.95 & 35.19 & 36.35 & 40.43 \\
         & GIAOTracker-Post & 40.72 & 27.10 & 57.46 & 36.65 & 36.72 & \textbf{45.69} \\
         & GIAOTracker-DetV2 & 43.12 & 37.44 & 58.95 & 39.93 & 35.31 & 43.95 \\
         & GIAOTracker-Fusion & \textbf{44.46} & \textbf{38.29} & \textbf{59.86} & \textbf{40.45} & \textbf{38.31} & 45.37 \\ \midrule[1pt]
         & GIAOTracker* & 82.97 & 77.69 & 79.72 & 71.15 & 73.33 & 91.67 \\
      \bottomrule[2pt]
      \end{tabular}
      \end{center}
      \caption{Overview of the results on test-dev dataset. "*" means taking annotation detections as input (best in bold).}
   \end{table*}

   \begin{table}[htbp]
      \begin{center}
      \begin{tabular}{cc|c}
      \toprule[2pt]
         & \textbf{Method}  &\textbf{\ \ \ mAP\ \ \ } \\ \midrule[1pt]
         & SOMOT & 58.61 \\  
         & \textbf{GIAOTracker (ours)} & \textbf{54.18} \\
         & MMDS & 52.68 \\
         & Deep IoU Tracker & 48.54 \\
         & Yolo-Deepsort-VisDrone & 46.70 \\
         & CenterPointCF & 44.03 \\
         & MIYoT & 39.35 \\
         & DeepTAMA+Homography+Voting & 39.25 \\
         & HNet & 24.71 \\
      \bottomrule[2pt]
      \end{tabular}
      \end{center}
      \caption{Top 9 results of the VisDrone2021 MOT Challenge \cite{leaderboard}.}
      \vspace{-1em}
   \end{table}

   \noindent \textbf{Overview.} Table 4 presents the overview results on VisDrone MOT test-dev dataset, which includes 16 sequences. 
   Taking DetV1+DeepSORT as baseline, the second to fourth rows respectively show the tracking performance after adding three stages, 
   i.e., online tracking, global link and post-processing (excluding fusion). 
   The fifth row is the results after replace DetV1 with DetV2, which has better detection performance (from 56.9 to 63.2 AP50). 
   After fusing the tracking results on the fourth and fifth row by our TrackNMS, we achieve 44.46 mAP (row 6). 
   Moreover, the last row shows the tracking performance by taking annotation detections as input of GIAOTracker, 
   which improves by 92\% compared to our best single model results on test-dev dataset. 
   It proves the effectiveness and potential of our GIAOTracker, whose tracking quality is greatly limited by detection accuracy.

\subsection{Comparison to State-of-the-art}

   Finally, we compare our GIAOTracker with state-of-the-art results of the VisDrone2021 MOT Challenge. 
   Table 5 lists the best 9 performing results and our GIAOTracker ranks 2nd, which proves the effectiveness of our framework. 
   Note that our detector is simply trained on VisDrone train+val dataset, whose performance is limited by the amount of data and detection methods 
   (the usage of extra data is allowed in the VisDrone2021 Challenge). 
   As mentioned in Sec 4.3, we argue that more accurate detection results would improve the performance significantly. 

\section{Conclusion}
   In this paper, we provide a comprehensive framework for multi-class multi-object tracking in drone videos.
   Inspired by the hierarchical data association strategy, it consists of three stages, i.e., online tracking, global link and post-processing.
   The online tracking generates reliable tracklets with information including camera movements, object appearance and object motion,
   then they are associated into trajectories according to tracklet features and spatio-temporal distances in the global link stage.
   The final post-processing stage refines the tracking results through four methods.
   Our GIAOTracker achieves the state-of-the-art results on the VisDrone MOT dataset and ranks 2nd in the VisDrone2021 MOT Challenge. 

\section{Acknowledgements}
   This work is supported by Chinese National Natural Science Foundation under Grants (62076033, U1931202).

{\small
\bibliographystyle{ieee_fullname}
\bibliography{egpaper_final}

\begin{thebibliography}{10}\itemsep=-1pt

\bibitem{visdrone}
\url{http://aiskyeye.com/}.

\bibitem{cofe}
\url{https://www.youtube.com/watch?v=iroNC_6cHLs&t=1s}.

\bibitem{yolov5}
\url{https://github.com/ultralytics/yolov5}.

\bibitem{lsvrc2017}
\url{http://image-net.org/challenges/LSVRC/2017}.

\bibitem{leaderboard}
\url{http://www.aiskyeye.com/leaderboard/}.

\bibitem{bergmann2019tracking}
Philipp Bergmann, Tim Meinhardt, and Laura Leal-Taixe.
\newblock Tracking without bells and whistles.
\newblock In {\em Proceedings of the IEEE/CVF International Conference on
  Computer Vision}, pages 941--951, 2019.

\bibitem{bewley2016simple}
Alex Bewley, Zongyuan Ge, Lionel Ott, Fabio Ramos, and Ben Upcroft.
\newblock Simple online and realtime tracking.
\newblock In {\em 2016 IEEE international conference on image processing
  (ICIP)}, pages 3464--3468. IEEE, 2016.

\bibitem{bochinski2017high}
Erik Bochinski, Volker Eiselein, and Thomas Sikora.
\newblock High-speed tracking-by-detection without using image information.
\newblock In {\em 2017 14th IEEE International Conference on Advanced Video and
  Signal Based Surveillance (AVSS)}, pages 1--6. IEEE, 2017.

\bibitem{bochinski2018extending}
Erik Bochinski, Tobias Senst, and Thomas Sikora.
\newblock Extending iou based multi-object tracking by visual information.
\newblock In {\em 2018 15th IEEE International Conference on Advanced Video and
  Signal Based Surveillance (AVSS)}, pages 1--6. IEEE, 2018.

\bibitem{bochkovskiy2020yolov4}
Alexey Bochkovskiy, Chien-Yao Wang, and Hong-Yuan~Mark Liao.
\newblock Yolov4: Optimal speed and accuracy of object detection.
\newblock {\em arXiv preprint arXiv:2004.10934}, 2020.

\bibitem{bodla2017soft}
Navaneeth Bodla, Bharat Singh, Rama Chellappa, and Larry~S Davis.
\newblock Soft-nms--improving object detection with one line of code.
\newblock In {\em Proceedings of the IEEE international conference on computer
  vision}, pages 5561--5569, 2017.

\bibitem{braso2020learning}
Guillem Bras{\'o} and Laura Leal-Taix{\'e}.
\newblock Learning a neural solver for multiple object tracking.
\newblock In {\em Proceedings of the IEEE/CVF Conference on Computer Vision and
  Pattern Recognition}, pages 6247--6257, 2020.

\bibitem{cai2018cascade}
Zhaowei Cai and Nuno Vasconcelos.
\newblock Cascade r-cnn: Delving into high quality object detection.
\newblock In {\em Proceedings of the IEEE conference on computer vision and
  pattern recognition}, pages 6154--6162, 2018.

\bibitem{chen2017enhancing}
Jiahui Chen, Hao Sheng, Yang Zhang, and Zhang Xiong.
\newblock Enhancing detection model for multiple hypothesis tracking.
\newblock In {\em Proceedings of the IEEE Conference on Computer Vision and
  Pattern Recognition Workshops}, pages 18--27, 2017.

\bibitem{chu2019famnet}
Peng Chu and Haibin Ling.
\newblock Famnet: Joint learning of feature, affinity and multi-dimensional
  assignment for online multiple object tracking.
\newblock In {\em Proceedings of the IEEE/CVF International Conference on
  Computer Vision}, pages 6172--6181, 2019.

\bibitem{chu2021spatial}
Peng Chu, Jiang Wang, Quanzeng You, Haibin Ling, and Zicheng Liu.
\newblock Spatial-temporal graph transformer for multiple object tracking.
\newblock {\em arXiv e-prints}, pages arXiv--2104, 2021.

\bibitem{dehghan2015target}
Afshin Dehghan, Yicong Tian, Philip~HS Torr, and Mubarak Shah.
\newblock Target identity-aware network flow for online multiple target
  tracking.
\newblock In {\em Proceedings of the IEEE Conference on Computer Vision and
  Pattern Recognition}, pages 1146--1154, 2015.

\bibitem{deng2009imagenet}
Jia Deng, Wei Dong, Richard Socher, Li-Jia Li, Kai Li, and Li Fei-Fei.
\newblock Imagenet: A large-scale hierarchical image database.
\newblock In {\em 2009 IEEE conference on computer vision and pattern
  recognition}, pages 248--255. Ieee, 2009.

\bibitem{dietterich2000ensemble}
Thomas~G Dietterich.
\newblock Ensemble methods in machine learning.
\newblock In {\em International workshop on multiple classifier systems}, pages
  1--15. Springer, 2000.

\bibitem{fagot2016improving}
Lo{\"\i}c Fagot-Bouquet, Romaric Audigier, Yoann Dhome, and Fr{\'e}d{\'e}ric
  Lerasle.
\newblock Improving multi-frame data association with sparse representations
  for robust near-online multi-object tracking.
\newblock In {\em European Conference on Computer Vision}, pages 774--790.
  Springer, 2016.

\bibitem{fan2020visdrone}
Heng Fan, Dawei Du, Longyin Wen, Pengfei Zhu, Qinghua Hu, Haibin Ling, Mubarak
  Shah, Junwen Pan, Arne Schumann, Bin Dong, et~al.
\newblock Visdrone-mot2020: The vision meets drone multiple object tracking
  challenge results.
\newblock In {\em European Conference on Computer Vision}, pages 713--727.
  Springer, 2020.

\bibitem{fang2018recurrent}
Kuan Fang, Yu Xiang, Xiaocheng Li, and Silvio Savarese.
\newblock Recurrent autoregressive networks for online multi-object tracking.
\newblock In {\em 2018 IEEE Winter Conference on Applications of Computer
  Vision (WACV)}, pages 466--475. IEEE, 2018.

\bibitem{feichtenhofer2017detect}
Christoph Feichtenhofer, Axel Pinz, and Andrew Zisserman.
\newblock Detect to track and track to detect.
\newblock In {\em Proceedings of the IEEE International Conference on Computer
  Vision}, pages 3038--3046, 2017.

\bibitem{fischler1981random}
Martin~A Fischler and Robert~C Bolles.
\newblock Random sample consensus: a paradigm for model fitting with
  applications to image analysis and automated cartography.
\newblock {\em Communications of the ACM}, 24(6):381--395, 1981.

\bibitem{gao2018revisiting}
Jiyang Gao and Ram Nevatia.
\newblock Revisiting temporal modeling for video-based person reid.
\newblock {\em arXiv preprint arXiv:1805.02104}, 2018.

\bibitem{girbau2021multiple}
Andreu Girbau, Xavier Gir{\'o}-i Nieto, Ignasi Rius, and Ferran Marqu{\'e}s.
\newblock Multiple object tracking with mixture density networks for trajectory
  estimation.
\newblock {\em arXiv preprint arXiv:2106.10950}, 2021.

\bibitem{girshick2015fast}
Ross Girshick.
\newblock Fast r-cnn.
\newblock In {\em Proceedings of the IEEE international conference on computer
  vision}, pages 1440--1448, 2015.

\bibitem{girshick2014rich}
Ross Girshick, Jeff Donahue, Trevor Darrell, and Jitendra Malik.
\newblock Rich feature hierarchies for accurate object detection and semantic
  segmentation.
\newblock In {\em Proceedings of the IEEE conference on computer vision and
  pattern recognition}, pages 580--587, 2014.

\bibitem{han2020mat}
Shoudong Han, Piao Huang, Hongwei Wang, En Yu, Donghaisheng Liu, Xiaofeng Pan,
  and Jun Zhao.
\newblock Mat: Motion-aware multi-object tracking.
\newblock {\em arXiv preprint arXiv:2009.04794}, 2020.

\bibitem{he2016deep}
Kaiming He, Xiangyu Zhang, Shaoqing Ren, and Jian Sun.
\newblock Deep residual learning for image recognition.
\newblock In {\em Proceedings of the IEEE conference on computer vision and
  pattern recognition}, pages 770--778, 2016.

\bibitem{henriques2014high}
Jo{\~a}o~F Henriques, Rui Caseiro, Pedro Martins, and Jorge Batista.
\newblock High-speed tracking with kernelized correlation filters.
\newblock {\em IEEE transactions on pattern analysis and machine intelligence},
  37(3):583--596, 2014.

\bibitem{hermans2017defense}
Alexander Hermans, Lucas Beyer, and Bastian Leibe.
\newblock In defense of the triplet loss for person re-identification.
\newblock {\em arXiv preprint arXiv:1703.07737}, 2017.

\bibitem{hornakova2020lifted}
Andrea Hornakova, Roberto Henschel, Bodo Rosenhahn, and Paul Swoboda.
\newblock Lifted disjoint paths with application in multiple object tracking.
\newblock In {\em International Conference on Machine Learning}, pages
  4364--4375. PMLR, 2020.

\bibitem{huang2008robust}
Chang Huang, Bo Wu, and Ramakant Nevatia.
\newblock Robust object tracking by hierarchical association of detection
  responses.
\newblock In {\em European Conference on Computer Vision}, pages 788--801.
  Springer, 2008.

\bibitem{kalal2010forward}
Zdenek Kalal, Krystian Mikolajczyk, and Jiri Matas.
\newblock Forward-backward error: Automatic detection of tracking failures.
\newblock In {\em 2010 20th international conference on pattern recognition},
  pages 2756--2759. IEEE, 2010.

\bibitem{kalman1960new}
Rudolph~Emil Kalman.
\newblock A new approach to linear filtering and prediction problems.
\newblock 1960.

\bibitem{keuper2018motion}
Margret Keuper, Siyu Tang, Bjoern Andres, Thomas Brox, and Bernt Schiele.
\newblock Motion segmentation \& multiple object tracking by correlation
  co-clustering.
\newblock {\em IEEE transactions on pattern analysis and machine intelligence},
  42(1):140--153, 2018.

\bibitem{kim2015multiple}
Chanho Kim, Fuxin Li, Arridhana Ciptadi, and James~M Rehg.
\newblock Multiple hypothesis tracking revisited.
\newblock In {\em Proceedings of the IEEE international conference on computer
  vision}, pages 4696--4704, 2015.

\bibitem{kim2016cdt}
Han-Ul Kim and Chang-Su Kim.
\newblock Cdt: Cooperative detection and tracking for tracing multiple objects
  in video sequences.
\newblock In {\em European Conference on Computer Vision}, pages 851--867.
  Springer, 2016.

\bibitem{kuhn1955hungarian}
Harold~W Kuhn.
\newblock The hungarian method for the assignment problem.
\newblock {\em Naval research logistics quarterly}, 2(1-2):83--97, 1955.

\bibitem{li2019global}
Jianing Li, Jingdong Wang, Qi Tian, Wen Gao, and Shiliang Zhang.
\newblock Global-local temporal representations for video person
  re-identification.
\newblock In {\em Proceedings of the IEEE/CVF International Conference on
  Computer Vision}, pages 3958--3967, 2019.

\bibitem{li2021semi}
Wei Li, Yuanjun Xiong, Shuo Yang, Mingze Xu, Yongxin Wang, and Wei Xia.
\newblock Semi-tcl: Semi-supervised track contrastive representation learning.
\newblock {\em arXiv preprint arXiv:2107.02396}, 2021.

\bibitem{lin2014microsoft}
Tsung-Yi Lin, Michael Maire, Serge Belongie, James Hays, Pietro Perona, Deva
  Ramanan, Piotr Doll{\'a}r, and C~Lawrence Zitnick.
\newblock Microsoft coco: Common objects in context.
\newblock In {\em European conference on computer vision}, pages 740--755.
  Springer, 2014.

\bibitem{milan2017online}
Anton Milan, S~Hamid Rezatofighi, Anthony Dick, Ian Reid, and Konrad Schindler.
\newblock Online multi-target tracking using recurrent neural networks.
\newblock In {\em Thirty-First AAAI Conference on Artificial Intelligence},
  2017.

\bibitem{milan2015multi}
Anton Milan, Konrad Schindler, and Stefan Roth.
\newblock Multi-target tracking by discrete-continuous energy minimization.
\newblock {\em IEEE transactions on pattern analysis and machine intelligence},
  38(10):2054--2068, 2015.

\bibitem{ondruska2016deep}
Peter Ondruska and Ingmar Posner.
\newblock Deep tracking: Seeing beyond seeing using recurrent neural networks.
\newblock In {\em Thirtieth AAAI conference on artificial intelligence}, 2016.

\bibitem{pan2019multi}
Siyang Pan, Zhihang Tong, Yanyun Zhao, Zhicheng Zhao, Fei Su, and Bojin Zhuang.
\newblock Multi-object tracking hierarchically in visual data taken from
  drones.
\newblock In {\em Proceedings of the IEEE/CVF International Conference on
  Computer Vision Workshops}, pages 0--0, 2019.

\bibitem{pang2020tubetk}
Bo Pang, Yizhuo Li, Yifan Zhang, Muchen Li, and Cewu Lu.
\newblock Tubetk: Adopting tubes to track multi-object in a one-step training
  model.
\newblock In {\em Proceedings of the IEEE/CVF Conference on Computer Vision and
  Pattern Recognition}, pages 6308--6318, 2020.

\bibitem{pang2021quasi}
Jiangmiao Pang, Linlu Qiu, Xia Li, Haofeng Chen, Qi Li, Trevor Darrell, and
  Fisher Yu.
\newblock Quasi-dense similarity learning for multiple object tracking.
\newblock In {\em Proceedings of the IEEE/CVF Conference on Computer Vision and
  Pattern Recognition}, pages 164--173, 2021.

\bibitem{peng2020chained}
Jinlong Peng, Changan Wang, Fangbin Wan, Yang Wu, Yabiao Wang, Ying Tai,
  Chengjie Wang, Jilin Li, Feiyue Huang, and Yanwei Fu.
\newblock Chained-tracker: Chaining paired attentive regression results for
  end-to-end joint multiple-object detection and tracking.
\newblock In {\em European Conference on Computer Vision}, pages 145--161.
  Springer, 2020.

\bibitem{peng2020tpm}
Jinlong Peng, Tao Wang, Weiyao Lin, Jian Wang, John See, Shilei Wen, and Erui
  Ding.
\newblock Tpm: Multiple object tracking with tracklet-plane matching.
\newblock {\em Pattern Recognition}, 107:107480, 2020.

\bibitem{pirsiavash2011globally}
Hamed Pirsiavash, Deva Ramanan, and Charless~C Fowlkes.
\newblock Globally-optimal greedy algorithms for tracking a variable number of
  objects.
\newblock In {\em CVPR 2011}, pages 1201--1208. IEEE, 2011.

\bibitem{qiao2021detectors}
Siyuan Qiao, Liang-Chieh Chen, and Alan Yuille.
\newblock Detectors: Detecting objects with recursive feature pyramid and
  switchable atrous convolution.
\newblock In {\em Proceedings of the IEEE/CVF Conference on Computer Vision and
  Pattern Recognition}, pages 10213--10224, 2021.

\bibitem{redmon2016you}
Joseph Redmon, Santosh Divvala, Ross Girshick, and Ali Farhadi.
\newblock You only look once: Unified, real-time object detection.
\newblock In {\em Proceedings of the IEEE conference on computer vision and
  pattern recognition}, pages 779--788, 2016.

\bibitem{redmon2017yolo9000}
Joseph Redmon and Ali Farhadi.
\newblock Yolo9000: better, faster, stronger.
\newblock In {\em Proceedings of the IEEE conference on computer vision and
  pattern recognition}, pages 7263--7271, 2017.

\bibitem{redmon2018yolov3}
Joseph Redmon and Ali Farhadi.
\newblock Yolov3: An incremental improvement.
\newblock {\em arXiv preprint arXiv:1804.02767}, 2018.

\bibitem{ren2015faster}
Shaoqing Ren, Kaiming He, Ross Girshick, and Jian Sun.
\newblock Faster r-cnn: Towards real-time object detection with region proposal
  networks.
\newblock {\em Advances in neural information processing systems}, 28:91--99,
  2015.

\bibitem{rublee2011orb}
Ethan Rublee, Vincent Rabaud, Kurt Konolige, and Gary Bradski.
\newblock Orb: An efficient alternative to sift or surf.
\newblock In {\em 2011 International conference on computer vision}, pages
  2564--2571. Ieee, 2011.

\bibitem{solovyev2021weighted}
Roman Solovyev, Weimin Wang, and Tatiana Gabruseva.
\newblock Weighted boxes fusion: Ensembling boxes from different object
  detection models.
\newblock {\em Image and Vision Computing}, 107:104117, 2021.

\bibitem{sun2020simultaneous}
ShiJie Sun, Naveed Akhtar, XiangYu Song, HuanSheng Song, Ajmal Mian, and
  Mubarak Shah.
\newblock Simultaneous detection and tracking with motion modelling for
  multiple object tracking.
\newblock In {\em European Conference on Computer Vision}, pages 626--643.
  Springer, 2020.

\bibitem{sun2018beyond}
Yifan Sun, Liang Zheng, Yi Yang, Qi Tian, and Shengjin Wang.
\newblock Beyond part models: Person retrieval with refined part pooling (and a
  strong convolutional baseline).
\newblock In {\em Proceedings of the European conference on computer vision
  (ECCV)}, pages 480--496, 2018.

\bibitem{vaswani2017attention}
Ashish Vaswani, Noam Shazeer, Niki Parmar, Jakob Uszkoreit, Llion Jones,
  Aidan~N Gomez, {\L}ukasz Kaiser, and Illia Polosukhin.
\newblock Attention is all you need.
\newblock In {\em Advances in neural information processing systems}, pages
  5998--6008, 2017.

\bibitem{voigtlaender2019mots}
Paul Voigtlaender, Michael Krause, Aljosa Osep, Jonathon Luiten, Berin
  Balachandar~Gnana Sekar, Andreas Geiger, and Bastian Leibe.
\newblock Mots: Multi-object tracking and segmentation.
\newblock In {\em Proceedings of the IEEE/CVF Conference on Computer Vision and
  Pattern Recognition}, pages 7942--7951, 2019.

\bibitem{wan2000unscented}
Eric~A Wan and Rudolph Van Der~Merwe.
\newblock The unscented kalman filter for nonlinear estimation.
\newblock In {\em Proceedings of the IEEE 2000 Adaptive Systems for Signal
  Processing, Communications, and Control Symposium (Cat. No. 00EX373)}, pages
  153--158. Ieee, 2000.

\bibitem{wang2021scaled}
Chien-Yao Wang, Alexey Bochkovskiy, and Hong-Yuan~Mark Liao.
\newblock Scaled-yolov4: Scaling cross stage partial network.
\newblock In {\em Proceedings of the IEEE/CVF Conference on Computer Vision and
  Pattern Recognition}, pages 13029--13038, 2021.

\bibitem{wang2021split}
Gaoang Wang, Yizhou Wang, Renshu Gu, Weijie Hu, and Jenq-Neng Hwang.
\newblock Split and connect: A universal tracklet booster for multi-object
  tracking.
\newblock {\em arXiv preprint arXiv:2105.02426}, 2021.

\bibitem{wang2019exploit}
Gaoang Wang, Yizhou Wang, Haotian Zhang, Renshu Gu, and Jenq-Neng Hwang.
\newblock Exploit the connectivity: Multi-object tracking with trackletnet.
\newblock In {\em Proceedings of the 27th ACM International Conference on
  Multimedia}, pages 482--490, 2019.

\bibitem{wang2015tracking}
Xinchao Wang, Engin T{\"u}retken, Francois Fleuret, and Pascal Fua.
\newblock Tracking interacting objects using intertwined flows.
\newblock {\em IEEE transactions on pattern analysis and machine intelligence},
  38(11):2312--2326, 2015.

\bibitem{wang2020towards}
Zhongdao Wang, Liang Zheng, Yixuan Liu, Yali Li, and Shengjin Wang.
\newblock Towards real-time multi-object tracking.
\newblock In {\em Computer Vision--ECCV 2020: 16th European Conference,
  Glasgow, UK, August 23--28, 2020, Proceedings, Part XI 16}, pages 107--122.
  Springer, 2020.

\bibitem{wen2019visdrone}
Longyin Wen, Pengfei Zhu, Dawei Du, Xiao Bian, Haibin Ling, Qinghua Hu, Jiayu
  Zheng, Tao Peng, Xinyao Wang, Yue Zhang, et~al.
\newblock Visdrone-mot2019: The vision meets drone multiple object tracking
  challenge results.
\newblock In {\em Proceedings of the IEEE/CVF International Conference on
  Computer Vision Workshops}, pages 0--0, 2019.

\bibitem{wojke2017simple}
Nicolai Wojke, Alex Bewley, and Dietrich Paulus.
\newblock Simple online and realtime tracking with a deep association metric.
\newblock In {\em 2017 IEEE international conference on image processing
  (ICIP)}, pages 3645--3649. IEEE, 2017.

\bibitem{wu2018exploit}
Yu Wu, Yutian Lin, Xuanyi Dong, Yan Yan, Wanli Ouyang, and Yi Yang.
\newblock Exploit the unknown gradually: One-shot video-based person
  re-identification by stepwise learning.
\newblock In {\em Proceedings of the IEEE conference on computer vision and
  pattern recognition}, pages 5177--5186, 2018.

\bibitem{xiang2015learning}
Yu Xiang, Alexandre Alahi, and Silvio Savarese.
\newblock Learning to track: Online multi-object tracking by decision making.
\newblock In {\em Proceedings of the IEEE international conference on computer
  vision}, pages 4705--4713, 2015.

\bibitem{xu2020train}
Yihong Xu, Aljosa Osep, Yutong Ban, Radu Horaud, Laura Leal-Taix{\'e}, and
  Xavier Alameda-Pineda.
\newblock How to train your deep multi-object tracker.
\newblock In {\em Proceedings of the IEEE/CVF Conference on Computer Vision and
  Pattern Recognition}, pages 6787--6796, 2020.

\bibitem{yang2012multi}
Bo Yang and Ram Nevatia.
\newblock Multi-target tracking by online learning of non-linear motion
  patterns and robust appearance models.
\newblock In {\em 2012 IEEE Conference on Computer Vision and Pattern
  Recognition}, pages 1918--1925. IEEE, 2012.

\bibitem{zamir2012gmcp}
Amir~Roshan Zamir, Afshin Dehghan, and Mubarak Shah.
\newblock Gmcp-tracker: Global multi-object tracking using generalized minimum
  clique graphs.
\newblock In {\em European conference on computer vision}, pages 343--356.
  Springer, 2012.

\bibitem{zhang2020multiple}
Jimuyang Zhang, Sanping Zhou, Xin Chang, Fangbin Wan, Jinjun Wang, Yang Wu, and
  Dong Huang.
\newblock Multiple object tracking by flowing and fusing.
\newblock {\em arXiv preprint arXiv:2001.11180}, 2020.

\bibitem{zhang2008global}
Li Zhang, Yuan Li, and Ramakant Nevatia.
\newblock Global data association for multi-object tracking using network
  flows.
\newblock In {\em 2008 IEEE Conference on Computer Vision and Pattern
  Recognition}, pages 1--8. IEEE, 2008.

\bibitem{zhang2020fairmot}
Yifu Zhang, Chunyu Wang, Xinggang Wang, Wenjun Zeng, and Wenyu Liu.
\newblock Fairmot: On the fairness of detection and re-identification in
  multiple object tracking.
\newblock {\em arXiv preprint arXiv:2004.01888}, 2020.

\bibitem{zheng2016mars}
Liang Zheng, Zhi Bie, Yifan Sun, Jingdong Wang, Chi Su, Shengjin Wang, and Qi
  Tian.
\newblock Mars: A video benchmark for large-scale person re-identification.
\newblock In {\em European Conference on Computer Vision}, pages 868--884.
  Springer, 2016.

\bibitem{zhou2019omni}
Kaiyang Zhou, Yongxin Yang, Andrea Cavallaro, and Tao Xiang.
\newblock Omni-scale feature learning for person re-identification.
\newblock In {\em Proceedings of the IEEE/CVF International Conference on
  Computer Vision}, pages 3702--3712, 2019.

\bibitem{zhou2020tracking}
Xingyi Zhou, Vladlen Koltun, and Philipp Kr{\"a}henb{\"u}hl.
\newblock Tracking objects as points.
\newblock In {\em European Conference on Computer Vision}, pages 474--490.
  Springer, 2020.

\bibitem{zhou2019objects}
Xingyi Zhou, Dequan Wang, and Philipp Kr{\"a}henb{\"u}hl.
\newblock Objects as points.
\newblock {\em arXiv preprint arXiv:1904.07850}, 2019.

\bibitem{zhu2018vision}
Pengfei Zhu, Longyin Wen, Xiao Bian, Haibin Ling, and Qinghua Hu.
\newblock Vision meets drones: A challenge.
\newblock {\em arXiv preprint arXiv:1804.07437}, 2018.

\bibitem{zhu2018visdrone}
Pengfei Zhu, Longyin Wen, Dawei Du, Xiao Bian, Haibin Ling, Qinghua Hu, Haotian
  Wu, Qinqin Nie, Hao Cheng, Chenfeng Liu, et~al.
\newblock Visdrone-vdt2018: The vision meets drone video detection and tracking
  challenge results.
\newblock In {\em Proceedings of the European Conference on Computer Vision
  (ECCV) Workshops}, pages 0--0, 2018.

\end{thebibliography}
}

\end{document}